# MixedPeds: Pedestrian Detection in Unannotated Videos using Synthetically Generated Human-agents for Training

**Ernest Cheung, Tsan Kwong Wong, Aniket Bera, Dinesh Manocha**
The University of North Carolina at Chapel Hill

## Abstract

We present a new method for training pedestrian detectors on an unannotated set of images. We produce a mixed reality dataset that is composed of real-world background images and synthetically generated static human-agents. Our approach is general, robust, and makes no other assumptions about the unannotated dataset regarding the number or location of pedestrians. We automatically extract from the dataset: i) the vanishing point to calibrate the virtual camera, and ii) the pedestrians' scales to generate a Spawn Probability Map, which is a novel concept that guides our algorithm to place the pedestrians at appropriate locations. After putting synthetic human-agents in the unannotated images, we use these augmented images to train a Pedestrian Detector, with the annotations generated along with the synthetic agents. We conducted our experiments using Faster R-CNN by comparing the detection results on the unannotated dataset performed by the detector trained using our approach and detectors trained with other manually labeled datasets. We showed that our approach improves the average precision by 5-13% over these detectors.

## Introduction

Accurate pedestrian detection is important for many autonomous systems, including self-driving cars, surveillance, robot navigation, etc. This problem has been extensively studied in computer vision, robotics and related areas. Recently, CNN-based pedestrian detectors have gained importance and been applied to different benchmarks (Dollár et al. 2012; 2009; Geiger, Lenz, and Urtasun 2012; Zhang et al. 2016; Cai et al. 2016; Jifeng Dai 2016).

The accuracy of CNN-based pedestrian detectors depends on the annotations in the training datasets. In order to achieve good accuracy, current methods ensure that the training data is from the same scene as or a similar environment to the testing data. These stipulations include similar camera configurations, lighting conditions, and backgrounds. This becomes an issue when one applies these methods to a new, unannotated video. In such cases, annotating training data can be challenging and requires considerable human effort. Overall, we need good pedestrian detection methods that can automatically work on unannotated videos.

One recent trend to generate labeled datasets is using synthetic and simulation approaches. In these methods, the pedestrian location and appearance in the image are generated using simulation techniques. With the pedestrian's location given, the resulting rendered image and computed annotations are used for training. Such techniques have also been used for pedestrian detection and crowd video analysis (Hattori et al. 2015; Cheung et al. 2016). However, current methods are either restricted to a fixed camera in terms of their usage or have a low detection accuracy. In this paper, we investigate new methods that can automatically generate annotated datasets for pedestrian detection by using simulation methods. The main motivation is to develop automated methods that can be used for a broad set of applications.

**Main Results:** We present a novel algorithm (MixedPeds) to generate the corresponding training data for CNN-based pedestrian detection, given an unannotated image dataset. Our approach combines real-world background information in a scene with synthetically-generated pedestrians, whose positions are precisely known. The annotated dataset is composed of synthetic pedestrians that can be controlled by different parameters. Our approach is applicable to videos captured from a moving camera and we present automatic feature extraction methods to obtain features that are used to place the synthetic agents. These novel techniques include computing the scale ratio, the vanishing point, the camera parameters, and a Spawn Probability Map. Furthermore, we present techniques to generate high-quality renderings and poses for synthetic pedestrians. As compared to prior method to perform pedestrian detection on an unannotated dataset, our approach offers the following benefits:

- A generalized framework to produce annotated data for training scene-specific pedestrian detectors.
- An automatic technique for feature extraction from images that are used to estimate the camera parameters and valid positions of pedestrians within an image.
- A data-driven approach to determinate lighting and clothes color of rendered pedestrians, to blend them into a real image better.
- An improvement in average precision of 7.8%, 5% and, 13.7% over prior methods on CALTECH, KITTI and ETHZ datasets, respectively.
- With few assumptions on the input dataset, the method can be applied on sets of images captured by any vehicle.

## Related Work

### Pedestrian Detection

Pedestrian detection is a sub-problem of object detection, which has been extensively studied in computer vision and related areas. Some of the earlier and popular methods are based on using HOG (Dalal and Triggs 2005) and SIFT (Lowe 2004) to extract features from images and using them to train different models (e.g. SVM (Suykens and Vandewalle 1999)). Inspired by a Convolution Neural Network (CNN) called AlexNet (Krizhevsky, Sutskever, and Hinton 2012), which demonstrated good results in terms of classifying objects, Girshick et al. (Girshick et al. 2014) proposed a method for transforming the object detection problem into a classification problem. This seminal method is known as Regions with CNN features (R-CNN).

Several methods (Girshick 2015; He et al. 2014; Ren et al. 2015) have been proposed to extract sub-region candidates to improve the efficiency and accuracy of R-CNN. Unlike prior methods, which use a selective search and spatial pyramid to generate sub-region candidates, Faster R-CNN also uses a deep network, namely Region Proposal Network (RPN), to compute the sub-region candidates. Several techniques have been proposed for accurate pedestrian detection like CALTECH(Dollár et al. 2012; 2009) and KITTI(Geiger, Lenz, and Urtasun 2012), which are based on Faster R-CNN. (Zhang et al. 2016) adopted the RPN in Faster R-CNN and combined it with a boosted forest for pedestrian detection. MS-CNN (Cai et al. 2016) proposed a similar network structure that has improvements for objects at a different scale. R-FCN (Jifeng Dai 2016) proposed a fully convolutional region-based detector that shows better efficiency. All these methods assume the availability of good annotated image datasets for training.

Other techniques for pedestrian detection are based on unsupervised learning (Sermanet et al. 2013; Benenson et al. 2014), though their accuracy tends to be lower than that of CNN-based pedestrian detectors.

### Synthetic Datasets for Machine Learning

With the demand for annotated data in the Deep Learning community, using a synthetic approach to produce data is becoming important to reduce manual annotation efforts. (Shrivastava et al. 2016) proposes an Adversarial Network approach to make synthetic data more realistic and preserve the annotation at the same time. (Varol et al. 2017) trains a CNN to learn from synthetic data and has shown its ability to perform human depth estimation and human part segmentation in real data. (Alexey Dosovitskiy 2017) launches an open framework to produce synthetic training data for autonomus driving.

Hattori et al. (Hattori et al. 2015) use synthetic datasets to improve pedestrian detection accuracy in a fixed camera video, where scene geometry and the camera perspective matrix are given. It overlays simulated pedestrians on the real-world background with no other pedestrians. This work has shown that simulated data can significantly improve the results in the static scene by providing a large number of training examples. Cheung et al. (Cheung et al. 2016) proposed a framework to generate a large amount of synthetic data for the pedestrians and the background objects in the scene.

### Camera Estimation and Pedestrian placement

To properly place pedestrians in an existing real-world image, the perspective information and the scene geometry information are needed. The perspective information can be obtained by camera calibration. However, many of the existing video datasets do not have camera parameters provided. (Caprile and Torre 1990), (Wang, Tsai, and others 1991) perform camera calibration using vanishing points and lines. In order to perform the computations automatically, these methods require automatic vanishing points/lines detection (Zhai, Workman, and Jacobs 2016; Nieto Doncel 2010). However, their performance and accuracy vary with different datasets.

Another set of works makes use of parallelepipeds (Wilczkowiak, Boyer, and Sturm 2001) and cuboids (Debevec, Taylor, and Malik 1996) to estimate the camera parameters, but they require manually selecting a few points. (Deutscher, Isard, and MacCormick 2006) uses perpendicular features in artificial information to automatically estimate the full camera model. However, this requires sufficient perpendicular features to work properly. Recent work in robotics on fully automatic calibration (Geiger et al. 2012; Levinson and Thrun 2013) relies on checkerboards (which may not be available in most unannotated datasets) or partially-known metric information (Yang et al. 2013).

Another problem that arises with synthetic methods is the computation of proper spawn locations. This leads to another well-studied problem in computer vision corresponding to scene segmentation. Despite the fact that there are effective existing methods (Kundu, Vineet, and Koltun 2016; Scharwächter et al. 2014; Ladický et al. 2012; Gould 2012) for handling this problem, they make certain assumptions and the results can vary with different datasets.

## Existing Approach for Pedestrian Detection on Unannotated Datasets

Given an unannotated dataset, we can apply a CNN-based algorithm trained on some other dataset to it. A key issue is to evaluate the accuracy of the pedestrian detection results, when the training and testing dataset are different. To evaluate the performance, we train pedestrian detectors using Faster R-CNN with ETHZ, Town Center, CALTECH, and KITTI training sets, and evaluate their accuracy. We show in the experiment section that the performance of the detector drops significantly when the training data and the testing data belong to different dataset(s).

## Methodology

In this section, we describe our approach to automatically generating an annotated training dataset using synthetic agents. We first describe our method of extracting two features: Pedestrian Scale Ratio and Vanishing Point, from the unannotated dataset. Then, we explain how we use them to estimate the camera parameters and determine the spawn locations of the synthetic agents, using a novel formulation: *Spawn Probability Map*. This approach is summarized in Algorithm 1. Finally, we describe our rendering techniques, which are used to compute the appearance of synthetic agents in the testing datasets.

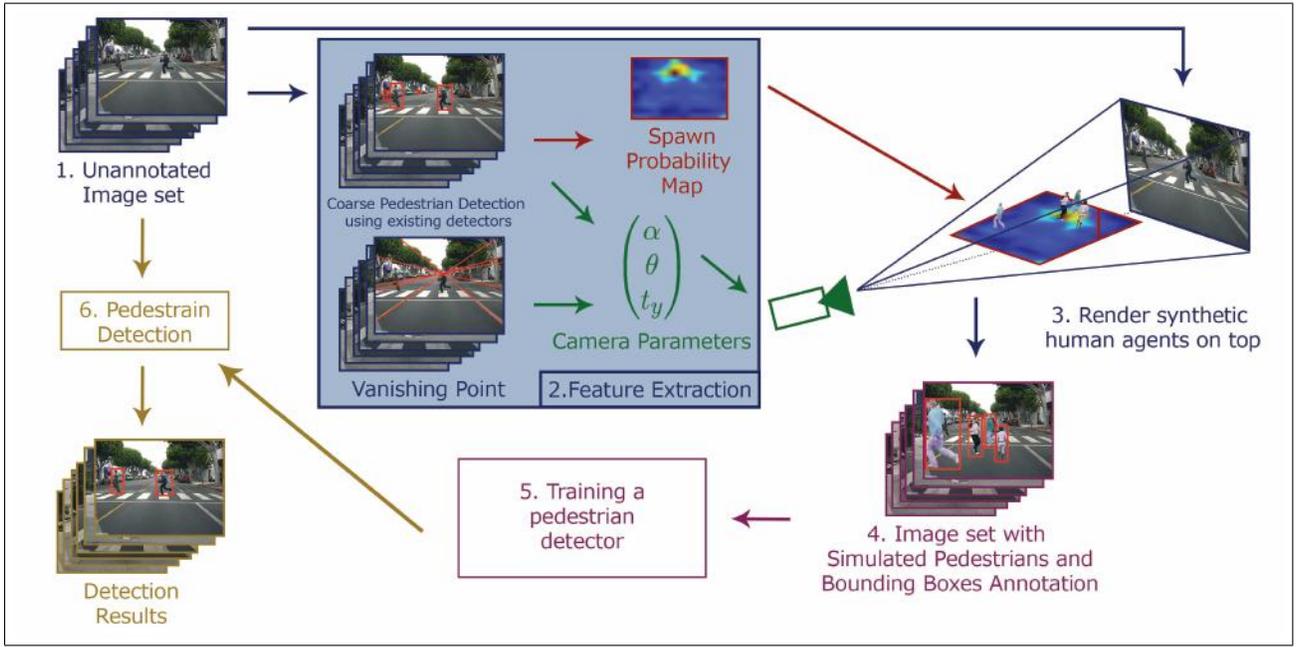

Figure 1: Overview: (1) We take an unannotated image set as input, and (2) extract the scale ratio of the pedestrian and vanishing points as features. Using these features, we compute the camera parameters along with a Spawn Probability Map that is used to determine the location of the synthetic agents. (3) We overlay synthetic pedestrians on the images to produce (4) an annotated training dataset, and (5) use it to train a pedestrian detector. (6) We perform pedestrian detection on the original dataset using this detector.

**Camera model** Our approach is designed for datasets captured by a camera mounted on a vehicle or a robot. Therefore, we can assume that its intrinsic matrix is the same throughout the dataset. For the extrinsic matrix, our assumption is that the camera is fixed on the vehicle and the vehicle is moving in a plane, which is almost parallel to the ground. Without loss of generality, we define the world coordinate frame as centered at the vertical projection of the camera on the ground, where the X axis points to the right, the Y-axis points forward towards the pedestrians, and the Z-axis points upward. We assume that the aspect ratio is one and there are no shear distortions in the camera. Moreover, we also assume that the camera is fixed upright on the vehicle such that there is no rotation about the optical axis (i.e. roll angle = 0) and the pitch angle is $\theta$. The projection matrix can be given as:

$$P = \begin{bmatrix} \alpha & 0 & u_0 \\ 0 & \alpha & v_0 \\ 0 & 0 & 1 \end{bmatrix} \begin{bmatrix} 1 & 0 & 0 & 0 \\ 0 & cos\theta & -sin\theta & t_y \\ 0 & sin\theta & cos\theta & 0 \end{bmatrix}. \quad (1)$$

Note that the principal point $(u_0, v_0)$ is given by half of the image width and height, respectively. Therefore, the parameters that are needed to estimate are $\alpha, \theta, t_y$, where $\alpha$ is the scaling factor of the camera, $\theta$ is the rotation angle about the X axis in the world coordinate frame, and $t_y$ is the distance of the camera from the ground. After obtaining these three parameters, we can estimate the Projection Matrix that can be applied to the entire dataset, and use these parameters for the virtual camera model that is used to overlay synthetic pedestrians. In the following part, we describe two features that are used to estimate the camera angle and techniques for extracting them and computing the camera matrix.

**Feature extraction** We extract two features from the unannotated dataset using a detector trained from another dataset. Our method is based on the following proposition:

**Proposition 1:** The ground truth values of the Scale of Pedestrian, Vanishing Point and Color of clothes of pedestrians within an image dataset can be estimated by the most confident results of a pedestrian detector that has poor average precision.

*Explanation:* Every single detected bounding has a confident score. Thus, if we only consider the most confident results (say top 10%), the overall precision is usually high ( 80 %) for detectors that has low average precision ( 30%). Therefore, we can make use of this set of confident bounding boxes, $BB_{top}$, to estimate ground truth of these features in the entire dataset, $BB_{gt}$. Experiment results conducted on KITTI and CALTECH dataset is consistent with this proposition.

**Scale Ratio:** With the same camera settings, the scale of pedestrians is the same for every image. We measure the scale of a pedestrian using the height of the bounding box in the image. Since the scale of the pedestrian varies depending on the distance from the camera, we take the scale ratio, $r$, as an invariant quantity that represents the scale of the pedestrian for the entire dataset. The scale ratio $r$ is defined as the ratio of the average height of the pedestrian bounding box $h$ to the vertical pixel coordinate of that pedestrian's foot $v_{foot}$. Note that having different heights of pedestrians in the data would introduce noise into our estimation, but we assume the effect of such noise would be nullified by sufficient data samples.

**Vanishing Point:** Since we assumed that the inclination between the vehicle and the ground is neglectable, the vanishing point $(u_{vanish}, v_{vanish})$ of all the images across the dataset will be close in numeric value.

To extract the scale ratio and vanishing point, we plot the $h$ and $v_{foot}$ for every pedestrian in $BB_{top}$ and use RANSAC(Fischler and Bolles 1981) to fit a line to this data. The slope of the line is the estimated scale ratio $r$ and the x-intercept is the estimated $v_{vanish}$ value, because the x-intercept of this plot represents the vertical pixel coordinate when the size of pedestrian bounding box vanishes to zero. We also show in the experiment Section that such estimation using only a small number of detections ($BB_{top}$ can precisely approximate the scale ratio and vanishing point of the ground truth data.

**Estimating Camera Parameters** Given equation 1, we can compute $h$ and $v_{foot}$ as follow. Let the world coordinate of the lowest point and highest point of a pedestrian be $(x, y_0, 0)^T$ and $(x, y_0, H)^T$, respectively. We denote $s\theta = sin\theta$ and $c\theta = cos\theta$. The height of the pedestrian in an image $h$ is given by:

$$h = \frac{(\alpha c\theta + v_0 s\theta)y_0 + \alpha t_y}{y_0 s\theta}$$
$$- \frac{(\alpha c\theta + v_0 s\theta)y_0 - (\alpha s\theta - v_0 c\theta)H + \alpha t_y}{y_0 s\theta + Hc\theta} \quad (2)$$
$$= \frac{\alpha H(y_0 + t_y c\theta)}{y_0 s\theta(y_0 s\theta + Hc\theta)}$$

Hence, the scale ratio can be expressed as:

$$r = \frac{h}{v_{foot}} = \frac{\alpha H(y_0 + t_y c\theta)}{((\alpha c\theta + v_0 s\theta)y_0 + \alpha t_y)} \quad (3)$$

The vanishing point is given by:

$$v_{vanish} = \lim_{Y \to \infty} \frac{(\alpha c\theta + v_0 s\theta)Y + \alpha t_y}{s\theta} = \frac{\alpha}{tan\theta} + v_0 \quad (4)$$

Using the $r$ and $v_{vanish}$, we estimate $\Theta = (\theta, \alpha, t_y)$ of the camera by performing optimization of this set of variables using gradient descent. We find a set of $\Theta$ within a reasonable parameter space from which $r$ and $v_{vanish}$ are extracted, such that the difference from the real camera matrix is minimized.

We compute optimal $\Theta$ by solving the equation below :

$$\arg\min_{\Theta} F(\Theta) = \arg\min_{\Theta} \lambda_1 \left\| (r' - r_\Theta) \right\|^2 + \lambda_2 \left\| (v'_{vanish} - v_\Theta) \right\|^2 \quad (5)$$

with lagrange multiplier $\lambda_1$, $\lambda_2$ where $r'$ and $v'_{vanish}$ are approximated by $BB_{top}$, with

$$\Theta^{j+1} = \Theta^j - \gamma \nabla F(\Theta) = \begin{bmatrix} \theta^j \\ \alpha^j \\ t_y^j \end{bmatrix} - \begin{bmatrix} \gamma \nabla_\theta F(\Theta) \\ \gamma \nabla_\alpha F(\Theta) \\ \gamma \nabla_{t_y} F(\Theta) \end{bmatrix} \quad (6)$$

Finally, we use the resulting $\Theta = (\theta, \alpha, t_y)$ to adjust the camera component in the Unreal Engine to estimate the Projection Matrix and overlay synthetic pedestrians.

### Spawn Probability Map

To avoid placing pedestrians in the wrong locations in the images, such as placing them on top of an obstacle, we present a data-driven approach to decide where to spawn pedestrians in the image: computing the Spawn Probability Map (SPM). We use the pedestrian detection results described above, $BB_{top}$, to construct a 2D histogram that indicates where the pedestrian can be spawned. Instead of merely incrementally the pixel in the histogram where the foot of the pedestrian is detected, we also incrementally increase the neighboring pixel using a Gaussian function. We repeat this process for every pedestrian detected in $BB_{top}$ to produce the SPM. When we spawn a synthetic agent in the image, we randomly decide the location base on the SPM: The higher the value of SPM, the higher the chance of spawning at that location.

In practice, it can be unrealistic to assume that the spawn location of the pedestrian follows the same SPM for every image in the dataset. This is because, when a vehicle is moving, different types of scenes can be captured: highway, crosswalk, tunnel, etc. Therefore, we apply K-means algorithm to cluster the images according to the average spawn location (i.e. average foot position of all detected pedestrian), and compute an SPM for each cluster. Since the distribution of pedestrian described by world coordinate is more meaningful than pixel coordinate, we perform a conversion using the estimated sets of camera parameters prior to clustering.

In some cases, the average spawn location can be sparsely distributed and hardly represent the location of the pedestrians within an image. We address this issue by first performing an initial clustering and measure the maximum distance in world coordinate, $d_{max}$, within each cluster. Taking average of all cluster, we define a metric, $d_{avg}/2$, to separate the images which pedestrians are sparsely distributed. These images are not considered during the clustering phases, and we compute another SPM for them to decide where to spawn the pedestrians.

### Rendering of Synthetic Agents

Generating high quality rendering of 3D models of synthetic agents can be very expensive. In this section, we describe how to solve the pedestrian placement issue and how to efficiently generate a set of reasonable color values for pedestrian appearance.

**Algorithm 1** Compute Spawn Probability Maps (SPMs)

**Input:** Unannotated Dataset $I$;
**Output:** A set of Spawn Probability Maps $SPM_c$; and Mappings for all images in $I$ to $SPM_c$;
1: Perform detection using existing detector on $I$;
2: Filter the $t\%$ most confident bounding boxes, $BB_{top}$;
3: Compute Scale Ratio $r$ and Vanishing Point $v_{vanish}$ from $BB_{top}$;
4: Estimate Camera parameters $P$ using $r$ and $v_{vanish}$
5: // Initial Clustering
6: **for all** image $I_k \in I$ **do**
7:     **for all** $BB_{top} \in I_k$ **do**
8:         $X_w = P(GetFootPixel(BB_{top}))$
9:     **end for**;
10:    Compute average location $A_k = avg(X_w)$ ;
11:    Compute $D_k = max(dist(X_w, X_w))$;
12: **end for**;
13: Cluster $I$ into $C_1...C_j$ using $A_k$
14: Compute $d_{max\_j}$ for $C_1...C_j$; $d_{avg} = max(d_{max\_j})$;
15: // Clustering for SPMs
16: $I_{sparse} = \{I_k \in I | D_k > d_{avg}/2\}$;
17: Repeat 6 to 12 with $I - I_{sparse}$ and compute new $A_k$ ;
18: Cluster $I - I_{sparse}$ into $C_1...C_j$ using new $A_k$;
19: Assign all images in $I_{sparse}$ to a seperate cluster $C_{j+1}$;
20: **for all** $C \in \{C_1, ..., C_{j+1}\}$ **do**
21:    **for all** $BB_{top} \in C$ **do**
22:        $SPM_j += N(GetFootPixel(BB_{top}), \sigma)$;
23:    **end for**;
24: **end for**;
25: **return** $(SPM_1...SPM_{j+1}, C_1...C_{j+1})$;

**3D Models of Synthetic Pedestrians** We use seven different 3D human models with variable clothing and skin color, and combine them with nine different poses to generate synthetic pedestrians. Skin color can be modeled based on the approaches described in (Vezhnevets, Sazonov, and Andreeva 2003). Unlike skin color, clothing colors in the real world tend to be more random. However, using a random color model can result in some unrealistic cloth appearances. For instance, some clothes color (pink, bright orange, and etc.) that we rarely find in the real world will be produced. Recall Proposition 1, we can use an HSV color model to generate clothing colors according to the color of the pedestrians we see in the unannotated dataset.

**Lighting Adjustment** Depending on the amount of lighting in the image, the brightness of the synthetic agent varies. If an agent is rendered under a shadow in the image, the agent should look darker, and if the agent is directly lit by the sunlight, it should look brighter. Therefore, we consider all the pixels in an image below the vanishing point and set a scale with the dimmest pixel and the brightest pixel. We map this scale to a reasonable range of values, which controls the brightness of the synthetic agents. For each synthetic agent in the resulting image, we consider the brightness of the neighboring pixels of the spawn coordinates and take the average to compute the brightness of the synthetic agents.

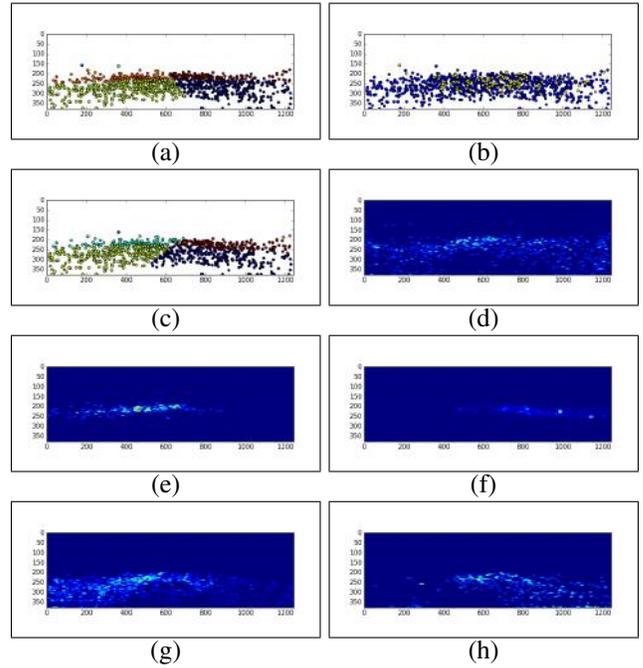

Figure 2: Clustering results [(c)] and Spawn Probability Maps [(d)-(h)] computed for KITTI dataset. (a) Initial clustering (step 6-12 in Algorithm 1). (b) Images with pedestrians sparsely distributed (yellow) are removed from $I - I_{sparse}$. (c) Clustering $I - I_{sparse}$ into 4 groups (step 16-18 in Algorithm 1). (d) SPM for $I_{sparse}(C_{j+1})$, which are used for images that contain sparsely distributed pedestrians. (e)-(h) SPMs for other clusters grouped in (c): $C_1...C_j$ (j=4).

## Experiments

### Feature Extraction

This experiment verifies Proposition 1 in CALTECH and KITTI dataset. We use a detector trained from the alternative dataset to produce the set of confident bounding box, $BB_{top}$, to estimate the Scale of Pedestrian, Vanishing Point and Color of clothes in the ground truth bounding boxes, $BB_{gt}$. Fig. 3 shows that estimation of Scale of Pedestrian($r$) and Vanishing Point ($v_{vanish}$), represented as the slope and x-intercept of the lines respectively, are close in value.

We compare the distribution of the HSV values of the ground truth pedestrians and the high-confidence pedestrians detected in KITTI dataset. The results are shwon in Fig. 4. As observed from Fig. 4, the distribution in the ground truth and high confidence detections are similar; pedestrians are more likely to have low saturation values, but random hue values. Therefore, we determine the hue and saturation according to the detection results, and determine the actual value based on the method described regarding Lighting adjustment.

### Pedestrian detection

We evaluated the results on three datasets that are captured by a moving vehicle: CALTECH(Dollár et al. 2012), KITTI(Geiger, Lenz, and Urtasun 2012), and ETHZ(Ess et al. 2009). We treat each datasets as an unannotated image sets when testing on it. The ground truth results provided with the dataset are used to evaluate the accuracy of CNN-based detection algorithm trained our annotated training dataset. We also validate our results using detectors trained from PASCAL-VOC(Everingham et al. 2015) and Town Center dataset(Benfold and Reid 2011) that are captured from alternative kind of cameras.

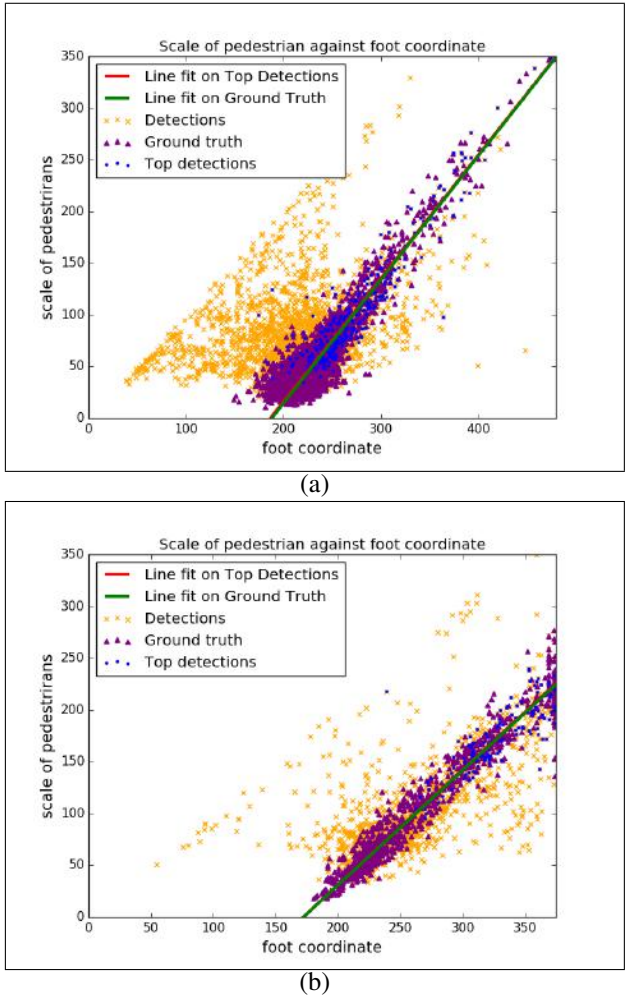

(a)

(b)

Figure 3: Extracting scale ratio and vanishing point by fitting a line to the plot of the Bounding Box height against the vertical coordinate of the pedestrian in Ground Truth (purple), detection (orange), and high confident detection $BB_{top}$ (blue). The slope and X-intercept of the line represent scale ratio and vanishing point, respectively. Results on both (a) CALTECH and (b) KITTI demonstrate that the line fitted on high confident detection (red) is almost overlapping with the line fitted on the ground truth data (green).

As discussed in previous section, several state-of-the-art pedestrian detectors are designed based on Faster R-CNN. Therefore, we also train Faster R-CNN pedestrian detectors to evaluate the benefits of our annotated training data. Noting that the choice of CNN used for pedestrian detection is orthogonal to our method, any pedestrian detector can be trained using our generated data.

We compare three variations of our approach with detectors trained from an alternative dataset, and show the precision-recall graph of the results in Fig. 5 and the average precision in Table 2.

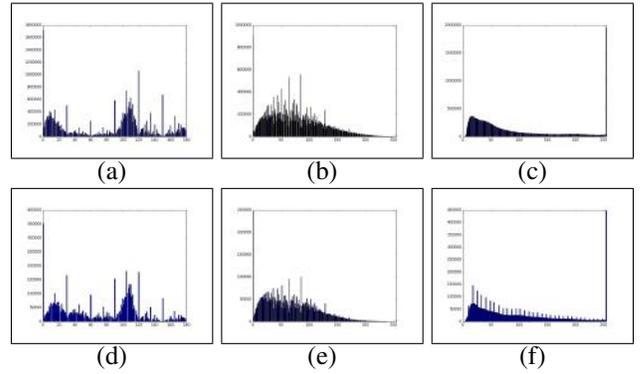

(a) (b) (c)

(d) (e) (f)

Figure 4: Color [Hue, Saturation, Value] histogram for pedestrians in ground truth, $BB_{gt}$ [(a),(b),(c)] and high-confidence detection results, $BB_{top}$ [(d),(e),(f)], respectively. Data distribution in ground truth and detection are found to be similar and thus the detected color histogram can be used to determine the color of the synthetic agents in our annotated dataset.

| Method | Training data of Faster R-CNN detector |
|---|---|
| MixedPeds | Our approach |
| MixedPeds* | Our approach without the rendering techniques |
| MixedPeds^ | Our approach without using SPMs (Pedestrian randomly spawn |
| ALL | Training Data from CALTECH, KITTI, ETHZ, TownCenter an |
| ALL* | Same as ALL, except the number of training samples drawn fro |

Table 1: Details of the variation of our approach and the detector trained with a combined dataset

| Train \ Test | CALTECH | KITTI | ETHZ |
|---|---|---|---|
| **MixedPeds** | **25.3%** (+7.8%) | **49.0%** (+5%) | **53.0%** (+13.7%) |
| MixedPeds* | 22.1% (+4.6%) | 45.4% (+1.4%) | 48.3% (+9%) |
| MixedPeds^ | 15.4% (-2.1%) | 31.0% (-13%) | 43.9% (+4.6%) |
| **ALL** | **17.5%** | **44.0%** | **39.3%** |
| ALL* | 16.2% | 38.2% | 31.6% |
| CALTECH | N/A | 27.8% | 6.52% |
| KITTI | 11.6% | N/A | 16.2% |
| ETHZ | 10.4% | 35.3% | N/A |
| TOWNC | 4.9% | 18.1% | 23.4% |
| PASCALVOC | 11.5% | 36.9% | 38.1% |

Table 2: The average precision of the detector results trained by our approach is compared against the detector trained using other datasets. The number in the brackets indicates the improvement of our approach over the detector trained with a combined dataset, ALL (See Table 1 for details), which is the best among all existing approaches.

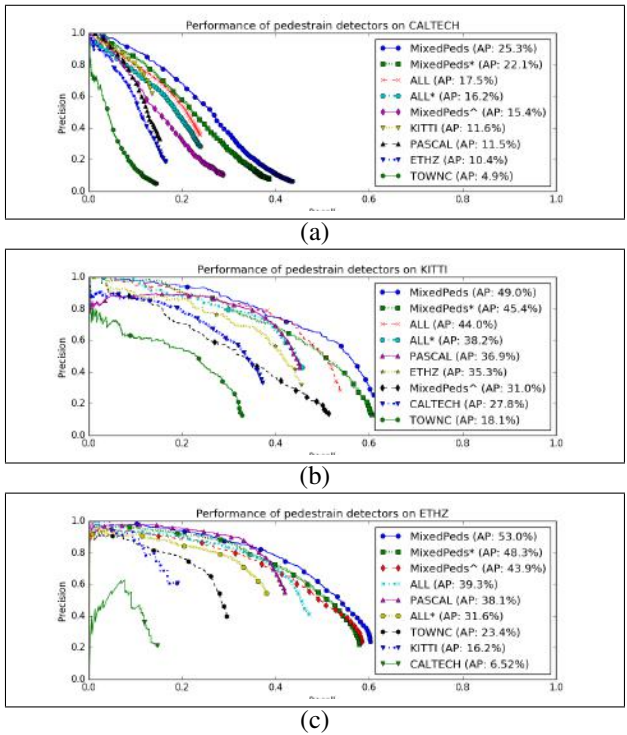

Figure 5: Precision-Recall graph of different detectors trained with the dataset and tested on (a) CALTECH, (b) KITTI, and (c) ETHZ. Our method exhibits better accuracy than existing detectors trained from other datasets. Also, MixedPeds* and MixedPeds^ demonstrate the benefits of Spawn Probability Maps and Rendering methods.

In the current approaches to detect pedestrians in unannotated dataset, combining training data from all existing dataset is the best. However, the detector trained with our approach is consistently better than the detector trained by these combined dataset. Besides, it also shows that the performance drops significantly when SPMs or our rendering techniques are not used in all three sets of experiments. This has shown the importance of using SPMs to determinate pedestrians' spawn location and applying our rendering technique. In terms of handling unannotated images, MixedPeds demonstrates significant benefits over prior methods. Fig. 6 shows some examples of the annotated datasets using our approach.

## Conclusion, Limitations, and Future Works

We present a new method to generate scene-specific training data from any unannotated dataset captured from the same camera that is fixed on a vehicle. Our method can be used to train pedestrian detectors that can considerably outperform other general-purpose pedestrian detectors by $5-13\%$. We also demonstrate the benefits of using synthetic datasets with appropriate rendering and spawning methods.

Our approach has some limitations. We assume that a majority of the images in the unannotated dataset are captured when the vehicle is moving on a plane (i.e. no inclination/declination). The performance of feature extraction can vary with the scenes and lighting conditions. As part of future work, we would like to explore robust automatic camera calibration methods for higher DOF cameras. It would be useful to incorporate segmentation algorithms to improve the positioning of spawning locations. We would also like to combine the recent approach (Shrivastava et al. 2016) to make our synthetic pedestrian look more realistic. Furthermore, we would also like to apply our method to train scene-specific pedestrian detectors for moving robots and vehicles.

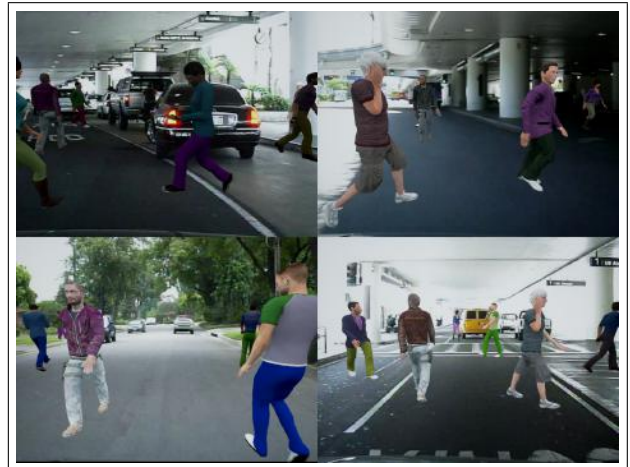

(a)

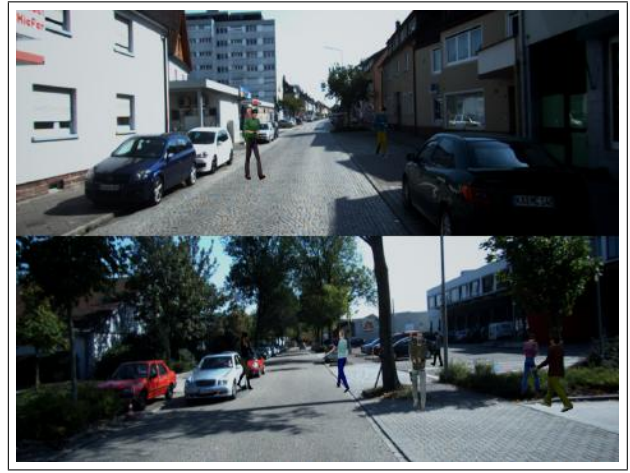

(b)

Figure 6: Example of augmented data. (a) CALTECH and (b) and KITTI. Synthetic agents look realistic with different appearances and brightness similar to the scene. For example, the pedestrian behind the car on the right in (b) upper image.